\def\year{2022}\relax
\documentclass[letterpaper]{article} 
\usepackage{aaai22}  
\usepackage{times}  
\usepackage{helvet}  
\usepackage{placeins}
\usepackage{amsfonts}
\usepackage{booktabs}
\usepackage{subfigure}
\usepackage{amsmath}
\usepackage{multirow}
\usepackage{amssymb}
\usepackage{courier}  
\usepackage[hyphens]{url}  
\usepackage{graphicx} 
\usepackage{xcolor}%

\usepackage{natbib}  
\usepackage{caption} 
\DeclareCaptionStyle{ruled}{labelfont=normalfont,labelsep=colon,strut=off} 
\usepackage{lipsum}

\renewcommand{\vec}[1]{\boldsymbol{#1}}
\newcommand{\mat}[1]{\boldsymbol{#1}}
\renewcommand{\mathbf}[1]{\boldsymbol{#1}}
\newcommand{\T}{^T}

\usepackage{algorithm}
\usepackage{algorithmic}

\usepackage{newfloat}
\usepackage{listings}
\floatstyle{ruled}
\newfloat{listing}{tb}{lst}{}
\floatname{listing}{Listing}
\title{Imputation of Missing Streamflow Data at Multiple Gauging Stations in Benin Republic}

\author {
    Rendani Mbuvha,\textsuperscript{\rm 1}\textsuperscript{\rm 2}
    Peniel Julien Yise Adounkpe, \textsuperscript{\rm 3}
    Wilson Tsakane Mongwe,\textsuperscript{\rm 4}
    Mandela Coovi Mahuwetin Houngnibo, \textsuperscript{\rm 5}
    Nathaniel Newlands,\textsuperscript{\rm 6}
    Tshilidzi Marwala \textsuperscript{\rm 4}
}
\affiliations {
    \textsuperscript{\rm 1} School of Electronic Engineering and Computer Science, Queen Mary University of London\\
    \textsuperscript{\rm 2} School of Statistics and Actuarial Science, University of Witwatersrand\\
    \textsuperscript{\rm 3} Institut National de l'Eau, Université d'Abomey-Calavi\\ 
    \textsuperscript{\rm 4}  School of Electrical and Electronic Engineering, University of Johannesburg\\
    \textsuperscript{\rm 5} Agence Nationale de la Météorologie du Bénin\\
    \textsuperscript{\rm 6} Summerland Research and Development Centre, Agriculture and Agri-Food Canada\\
    r.mbuvha@qmul.ac.uk, adounkpep@gmail.com, wilsonmongwe@gmail.com, hmandelahmadiba@gmail.com, nathaniel.newlands@agr.gc.ca, tmarwala@uj.ac.za
}

\usepackage{bibentry}

\begin{document}

\maketitle

\begin{abstract}

Streamflow observation data is vital for flood monitoring, agricultural, and settlement planning. However, such streamflow data are commonly plagued with missing observations due to various causes such as harsh environmental conditions and constrained operational resources. This problem is often more pervasive in under-resourced areas such as Sub-Saharan Africa. In this work, we reconstruct streamflow time series data through bias correction of the GEOGloWS ECMWF streamflow service (GESS) forecasts at ten river gauging stations in Benin Republic. We perform bias correction by fitting Quantile Mapping, Gaussian Process, and Elastic Net regression in a constrained training period. We show by simulating missingness in a testing period that GESS forecasts have a significant bias that results in low predictive skill over the ten Beninese stations. Our findings suggest that overall bias correction by Elastic Net and Gaussian Process regression achieves superior skill relative to traditional imputation by Random Forest, k-Nearest Neighbour, and GESS lookup. The findings of this work provide a basis for integrating global GESS streamflow data into operational early-warning decision-making systems (e.g., flood alert) in countries vulnerable to drought and flooding due to extreme weather events.
\end{abstract}

\section{Introduction}
Discharge predictions are critical for decision-making in many areas of climate change adaptation, including flood and drought prevention, agricultural planning, and hydroelectric power system operations \citep{kratzert2019neuralhydrology}. Predictions can 
be generated by physical and statistical models of the hydrological process within the river basin~\citep{adounkpepredicting}. Regardless of the approach, observation data is required to sufficiently calibrate the localised dynamics within catchments. Model-based predictions are even more essential in vulnerable and under-developed regions that are less resilient to extreme weather events such as drought and flooding. However, localised observations for hydrological models are not always available due to physical sensor failure events \citep{hamzah2020imputation}. Physical sensor systems often endure harsh environmental conditions, destruction and power outages \citep{tencaliec2015reconstruction}. These phenomena result in prolonged periods of missing or invalid observation readings. In developing regions such as Benin Republic, the effect of such missing data is further complemented by the already sparse distribution of stream gauging stations \citep{harrigan2020glofas}. Fitting statistical models based on incomplete observation data results in inaccurate models that do not capture the full dynamics of the underlying hydrological processes \citep{hamzah2020imputation}. A more complete set of streamflow data allows for greater confidence in the outputs of hydrological models \citep{pigott2001review}. Impuation of missing data therefore leads to much needed improved flood and drought predictions in a climate change context where these events are expected to increase in intensity and frequency \citep{kalantari2018nature}.  Thus missing data imputation has become a necessary first step in data preprocessing tasks \citep{pigott2001review}. Moreover, missing observations and the estimation of streamflow at ungauged locations are two of the most pervasive problems in the field of hydrology \citep{hamzah2020imputation}. Since the area of missing data imputation commands a relatively vast amount of statistical literature, a wide variety of techniques have been developed and applied with different data requirements and accuracy ~\citep{little2019statistical}. 

A simple approach to handling missing data is discarding records where any missingness arises. \citet{gill2007effect} show that while deletion of missing data is common practice in hydrology, it is not necessarily optimal as significant amounts of predictive information can be lost. On the other hand, single imputation methods attempt to replace missing values one feature at a time with a pre-specified summary statistic (e.g. mean or median) of the complete data \cite{baraldi2010introduction}. Regression-based methods such as imputation by Random Forest, k-Nearest Neighbours and Multilayer perceptrons  improve upon single imputation by exploiting correlations between variables in complete data \citep{gao2017dealing,pantanowitz2009missing}. Both regression and single imputation can lead to biases in the resultant analysis when the assumption that data are missing at random is violated \cite{gao2017dealing}. Multiple imputation methods such as Multivariate Imputation by Chained Equations (MICE) reduce this bias by replacing missing data points with predictions from an ensemble of regressors obtained from multiple subsets of the complete data \citep{van2018flexible}. 

Each of the imputation techniques above is used extensively in hydrology. \citet{hamzah2021comparison} compares numerous regression methods to reconstruct streamflow data in the Malaysian Langat River Basin. \citet{norazizi2019comparison} employs MICE and expectation maximisation (EM) for the imputation of rainfall data. \citet{adeloye2012self} and \citet{mwale2012infilling} propose the use of self-organising maps (SOMs) for infilling streamflow data at inadequately gauged basins in South West Nigeria and the Shine river basin in Malawi. In this work, we propose further augmenting regression imputation with data from Global streamflow forecasting systems.

Global streamflow prediction systems present a potential solution to missing data and the sparse distribution of stream gauge stations in developing countries. The Global Flood Awareness System (GLoFAS) is one such system jointly developed by the European Commission's Joint Research Center (JRC) and the European Centre for Medium-Range Weather Forecasting (ECMWF) \cite{Harrigan_2020}. GloFAS transforms numerical weather predictions to surface runoff through the hydrology tiled ECMWF scheme for surface exchanges over land (HTESSEL) model \citep{sanchez2021streamflow}. These surface runoff estimates are then routed to stream networks using the LISFOOD river routing scheme \citep{snow2015new, Harrigan_2020}. A limitation of the GloFAS system is that streamflow estimates are only available at a low resolution of a 0.1-degree grid cell (10 km$^2$); this limits the applicability of GloFAS to mainly large watersheds \cite{snow2015new,souffront2019hydrologic} . 

The Group on Earth Observations Global Water and Sustainability Initiative (GEOGloWS) attempts to address the coarse resolution of GloFAS with the GEOGloWS ECMWF streamflow service (GESS) \cite{ashby2021hydroviewer}. The GESS extends GloFAS to local sites using an updated digital elevation model (DEM) that allows for increased resolutions in areas of the world with narrow watersheds \citep{ashby2021hydroviewer}. Similar to GloFAS, runoff predictions from the HTESSEL land surface model are further downscaled and routed through a high-resolution stream network using the Muskingum algorithm of the routing application for parallel-discharge computation (RAPID) \citep{gill1978flood}.  

While GloFAS and GESS provide much-needed streamflow data, localised calibration is still a significant challenge. A global calibration of these systems is biased primarily to developed regions with dense and complete observations data \cite{ashby2021hydroviewer}. Thus, the utility of streamflow forecasts from these systems can be significantly enhanced by bias and variance correction to minimise systemic biases between model output and gauging station observations. Statistical Learning methods are broadly popular when learning transfer functions between global model forecasts and local observations \citep{PIANI2010199,rischard2018bias,wang2021use}. Various statistical learning methods have been put forward for bias correction tasks, including penalised regression, Quantile Mapping, Artificial Neural Networks and Gaussian Processes \citep{xu2021machine,pasten2020evaluation,hess-2022-53}.  

This paper presents and evaluates the bias correction of GESS river discharge estimates using statistical learning methods on ten hydrological gauging stations in Benin Republic. The proposed bias correction methods can be utilised to impute missing observation data with higher accuracy than standalone GESS and are superior to complete data-based Random Forest and k-Nearest Neighbour imputation.

\begin{figure}[ht]
\centering
\includegraphics[width=\columnwidth]{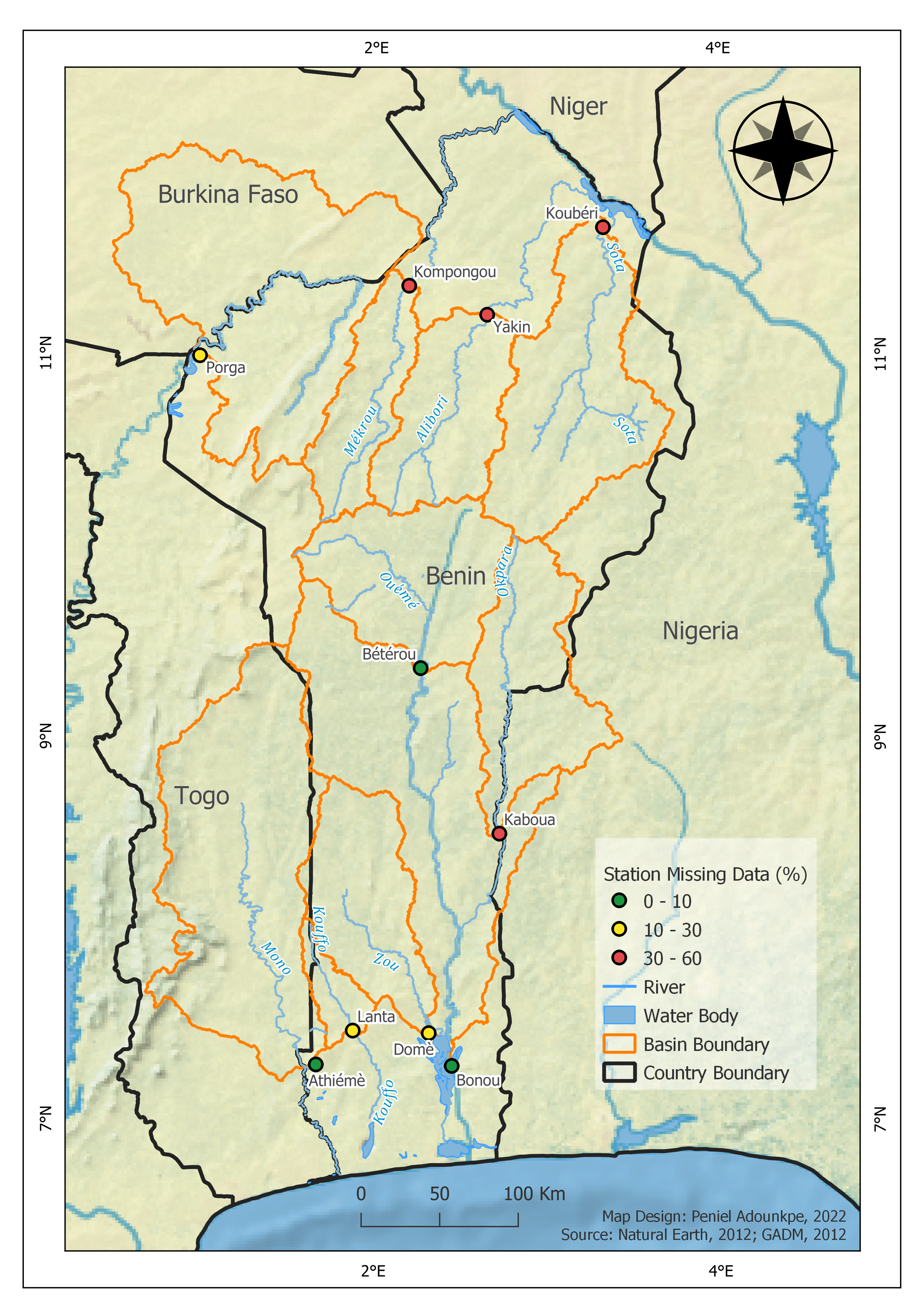}
\caption{Benin's catchments, rivers and hydrological stations missing data rate.}
\label{fig1}
\end{figure}

\section{Methods}

\begin{figure*}[ht]
\centering
\includegraphics[width=2\columnwidth]{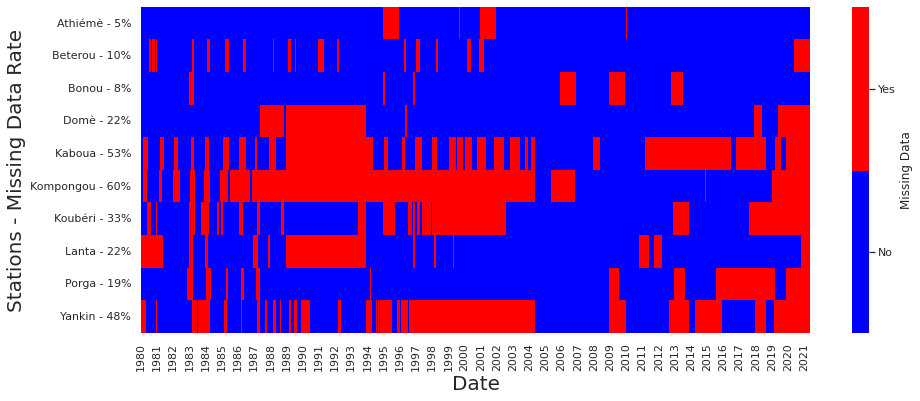}
\caption{Missing data periods at each hydrological gauging station.}
\label{fig2}
\end{figure*}
\subsection{Study Area}

For the purposes of this study, ten hydrological stations located on the outlets of Benin's major river basins were selected: Athiémè, Beterou, Bonou, Domè, Kaboua, Kompongou, Koubéri, Lanta, Porga and Yakin. Benin's basins are not limited to only Benin's administrative boundary but extend to its neighbouring countries: Togo, Burkina Faso and Nigeria. The Kouffo, Mono, Okpara, Ouémé and Zou rivers originate from central Benin and Togo and generally flow into the Atlantic Ocean. The Alibori, Mékrou and Sota rivers also originate from the centre of Benin but have their outlet up north in the Niger river. The Pendjari river (sometimes referred to as the Oti river) originates from Northern Benin and flows into the Volta river in Togo and Ghana.

\subsection{Data}
{\it In-situ} hydrological data (daily river discharge record at ten locations from 1980 to 2021) was acquired from Benin's hydrological service,  Direction Générale de l'Eau (DG-Eau). The percentage of missing data for the selected hydrological stations is 27.9\%, with the station of Athiémè having the least and the Kompongou station having the most. Figure \ref{fig2} shows in detail the periods of missing data for each station. On the other hand, the GESS forecasts present no missing data and provide daily river discharge data ranging from 1979 to the present at targeted river sections \cite{ashby2021hydroviewer}.

\begin{figure*}[h]
    \centering
    \includegraphics[width=0.9\textwidth]{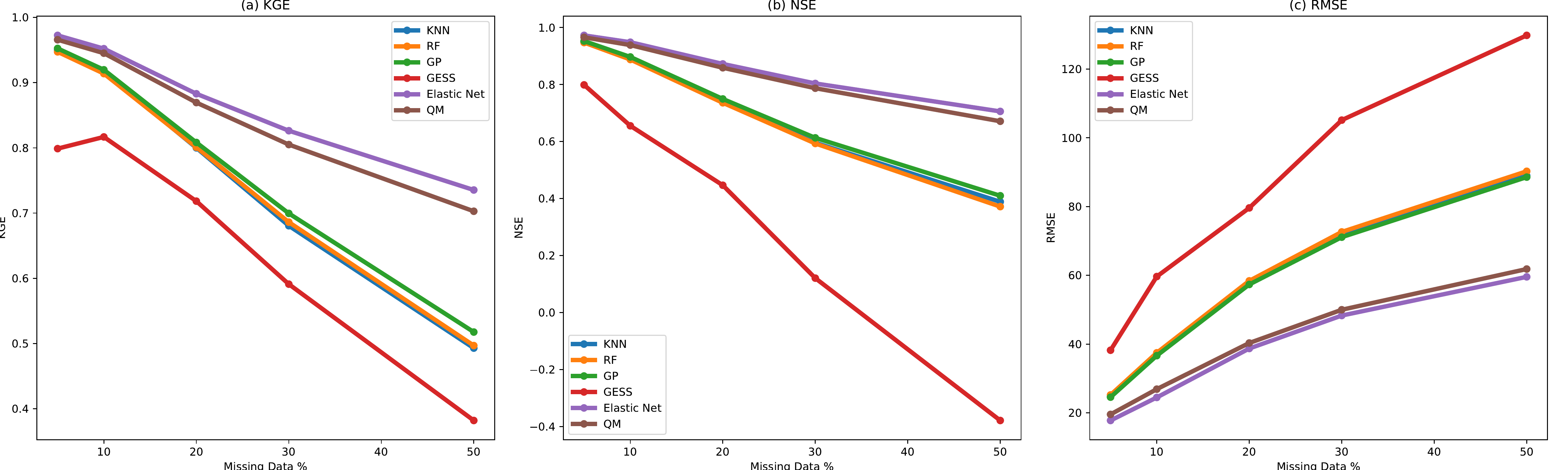}
    \caption{Plot of Mean KGE, NSE and RMSE measures from each imputation method at varying levels of missingness across the ten stations. }
    \label{fig:kge_nse_plt}
\end{figure*}

\subsection{Imputation Methods}

We investigate streamflow imputation by bias-correcting GESS forecasts. We propose an imputation scheme where we replace a missing {\it in-situ} observation at time $t$ with:
\begin{equation}
    \Tilde{\vec x_t} = h(\vec x_t^\prime)
\end{equation}
where $\vec x^\prime$  the GESS forecast of streamflow at time $t$. $h$ is a regressor trained on periods where $\vec x$ and  $\vec x^\prime$ are both available. We investigate settings where $h$ takes various forms, including; a simple lookup from GESS, a quantile mapping,  Elastic Net and Gaussian Process.

\subsubsection{Gaussian processes (GPs)} are nonparametric probabilistic  models, that are well known for regression problems \cite{cohen2020healing}. A GP is a collection of random variables, where each of its finite subsets is jointly Gaussian \cite{Rasmussen:2005:GPM:1162254}. GPs are sufficiently defined by a mean function $\mu(\vec x^\prime)$ and a kernel function $k(\vec x_i^\prime,\vec x_j^\prime)$. 

Given a training dataset $\{ \mathbf{x}^\prime_t,\mathbf{x}_t\}^t_{t=1}$ with $t$ noisy observations $\mathbf{x}_t=h(\vec x^\prime_t)+\epsilon$, where $\epsilon \sim\mathcal{N}(0,\sigma_X^2)$. The GP prior on $h$  is such that $h(\vec x^\prime) \sim \mathcal N(\vec \mu_{x^\prime},\mat K_{x^\prime}+\sigma_{X}^2\mat{I})$. The mean and variance of the Gaussian posterior predictive distribution of the function $h(\vec{x^\prime}_*)$ at a test point $\vec x_*^\prime$, are given by
\begin{align}
\mathbb{E}(h_*)& = \mathbf{\mu_{x^\prime_*}}+\mat k_*\T(\mat K_{x^\prime}+\sigma_{X^\prime}^2\mathbf{I})^{-1}(\mathbf{X}-\mathbf{\mu_{x^\prime})}, 
\label{eq:gppos mean}
\\
\text{var}(h_*)& = \mat k_{**}-\mathbf{k}_*^{T}(\mat K_{x^\prime}+\sigma_{X^\prime}^2\mathbf{I})^{-1}\mathbf{k}_*
\label{eq:gppos var}
\end{align}
where $\mat k_{**}=k(\mathbf{x^\prime}_*,\mathbf{x^\prime}_*)$ and  $\mat k_*=k(\mathbf{x^\prime},\mathbf{x^\prime}_*)$. 
The kernel hyperparameters $\vec \theta$ and the noise parameter $\sigma_{\vec X}$ are obtained by maximising the log-marginal likelihood
\begin{equation}
\label{eq:gplikelihood}
    \log p(\vec X|\mat X^\prime) =\log \mathcal{N}(\vec \mu_{x^\prime},\mat K_{x^\prime x^\prime}+ \sigma_{\vec X}^2\mathbf{I}\big).
\end{equation}
In this work, we use a multi-output formulation of the GP where all ten stations share a squared exponential kernel, thus allowing for implicit inference of connectivity relationships between stations.

\subsubsection{Elastic Net}
The elastic net is a penalised linear regression model that employs weighted $l_1$ and $l_2$  norm regularisation terms. The regularisation terms serve as conservative priors (bias towards zero) on the coefficients that prevent over-fitting to training data. Similar to the GP, we utilised a multi-output formulation of the Elastic Net that allows information sharing between stations. In our imputation formulation;
\begin{equation}
 \Tilde{\vec x}_t=h(\vec x^\prime_t) = \vec {X}^\prime \vec \beta +\epsilon
\end{equation}
where $\vec\beta$ is a matrix of coefficients on the GESS predictions $\vec {x}^\prime_t$. The matrix of coefficients is obtained by optimising the posterior : 
\begin{equation}
   \|\vec X- \vec X^\prime\vec\beta \|^{2}+\lambda _{2}\|\vec\beta \|^{2}+\lambda _{1}\|\vec\beta \|_{1}
\end{equation}
with the hyperparametes $\lambda_1$ and $\lambda_2$ obtained by cross validation with the constraint $\lambda_1 +\lambda_2 = 1$. Given limited in-situ data, the elastic net has the advantage of limiting over-fitting the bias correction relative to a regularised regression.
\subsubsection{Quantile Mapping}
 Quantile Mapping (QM) is a well-known method for bias correction of physical model output \citep{maraun2013bias,ringard2017quantile}. In QM, we seek to learn a mapping between the empirical cumulative distribution functions (CDF) of the in-situ and GESS prediction. In this case, the bias correction transfer function is: 
\begin{equation}
    \Tilde{\vec x}_t=h(\vec x^\prime_t) = F_X^{-1}F_{X^\prime}(x^\prime_t)
\end{equation}
Where $F_X^{-1}$ is the inverse empirical CDF of the in-situ data and $F_{X^\prime}$ is the empirical CDF of the GESS data. Both empirical CDFs are obtained during a specified training period. 
\subsubsection{ Baselines}
We compare imputation by bias correction using the abovementioned methods to traditional regression imputation methods that rely only on complete in-situ data. We consider regression-based imputation by Random Forest (RF) \cite{pantanowitz2009missing} and  k-nearest neighbours \citet{hamzah2021comparison} as our baselines.

\subsection{Performance Evaluation}

We measure the quality of the imputation using Kling-Gupta Efficiency (KGE) \cite{GUPTA200980}, Nash-Sutcliffe Efficiency (NSE) \cite{NASH1970282} and Root Mean Square Error (RMSE) \cite{gmd-7-1247-2014}. 

The KGE metric provides a diagnostically informative decomposition of the NSE and RMSE. It facilitates the analysis of the relative importance of its different components (correlation, bias and variability) in the context of hydrological modelling. The NSE measures the ability to predict variables different from the mean and gives the proportion of the initial variance accounted for by the model. The RMSE is frequently used to evaluate how closely the predicted values match the observed values based on the relative range of the data.

\begin{equation}
\centering
\begin{aligned}
    KGE = 1 - \sqrt{ (r -1) + \left(\dfrac{\bar{\hat{x_t}}}{\bar{x_t}} - 1\right)^2 +  \left(\dfrac{\sigma_{\hat{x_t}}}{\sigma_{x_t}} - 1\right)^2}
\end{aligned}
\end{equation}

\begin{equation}
\centering
\begin{aligned}
    NSE = 1 - \dfrac{\sum\limits_{i=1}^{n}(x_t-\hat{x_t})^2}{\sum\limits_{i=1}^{n}(x_t-\bar{x_t} )^2}
\end{aligned}
\end{equation}

\begin{equation}
\centering
\begin{aligned}
    RMSE = \sqrt{ \dfrac{1}{n} \sum_{i=1}^{n} \left(x_t - \hat{x_t} \right)^2}
\end{aligned}
\end{equation}

In the above equations, $x_t$ represents the observed discharge at a time $t$, $\hat{x_t}$ represents the imputed discharge at a time $t$, $\bar{x_t}$ is the mean of the observed discharge, $\bar{\hat{x_t}}$ is the mean of the imputed discharge, $\sigma_{\hat{x_t}}$ is the variance of the simulated discharge, $\sigma_{\hat{x_t}}$ is the variance of the observed discharge, $r$ is the Pearson correlation coefficient, and $n$ is the number of observations. The KGE values range from $-\infty$ to 1, with values close to 1 indicating good agreement. Similarly to the KGE, the NSE values also range from $-\infty$ to 1, with the perfect model having the value 1. The RMSE values range from 0 to $\infty$, with 0 implying the model has a perfect fit.

\section{Results and Discussion}

We partition the data such that 60\% of the period (1980/01-2004/11) is for training and 40\% is for (2004/11-2021/06) testing. We evaluate the performance of the imputation methods by randomly simulating missingness on the complete testing data at rates of  5\%, 10\%, 20\%, 30\% and 50\%.

\par Figure \ref{fig:kge_nse_plt} shows the mean KGE, NSE and RMSE of the imputation methods across gauging stations and levels of missingness. It can be seen that a simple lookup of the GESS predictions yields the worst imputation performance. The low KGE  and high RMSE values detailed in Table \ref{tab:results} suggest that the GESS predictions have significant bias. 

\par The hypothesis around significant bias in the GESS predictions is reinforced by the outperformance of the GP and Elastic Net bias correction imputation. Importantly the imputation by bias correction yields better performance than simply employing RF, KNN and MLP regressors on the complete in-situ data. This can be reasonably expected as there are limited periods where the data is jointly complete for training the multiple imputation regressors. Thus the GESS data provides the necessary signals regarding discharge seasonality and hydrologic connectivity to enhance imputation quality significantly.

\begin{table}
    \centering
\scalebox{0.5}{    
\begin{tabular}{lccccc|ccccc }
 
\toprule
\multirow{3}{*}{Method}&\multicolumn{5}{c}{NSE}&\multicolumn{5}{c}{RMSE} \\
&\multicolumn{10}{c}{\% Missing Data} \\\cline{2-11}\\

 & 5\% & 10\%& 20\% &30\% &50\%&5\% & 10\%& 20\% &30\% &50\% \\
\midrule
KNN &0.950&	0.892&	0.739&	0.596&  0.389&	 24.694 &	 36.802 &	 57.686 &	 71.743 &	 88.948\\ 
RF & 0.947&	0.888&	0.736&	0.593&	0.372&	 25.222 &	 37.476 &	 58.400 &	 72.625 &	 90.265 \\
Gaussian Process & \textbf{0.973}&	\textbf{0.938}&	0.840&	0.751&	0.623&	 \textbf{17.998} &	 28.308 &	 45.950 &	 57.364 &	 71.025 \\
GESS &0.799&	0.655&	0.447&	0.121&	-0.378&	 38.219 &	 59.650 &	 79.612 &	 105.145 &	 129.798 \\
Elastic Net& \textbf{0.973}&	\textbf{0.948}&	\textbf{0.872}&	\textbf{0.804}&	\textbf{0.706}&	 \textbf{17.783} &	 \textbf{24.484} &	 \textbf{38.719} &	 \textbf{48.309} &	 \textbf{59.559} \\
Quantile Mapping & 0.966&	0.938&	0.859&	0.787&	0.671&	 19.557 &	 26.900 &	 40.323 &	 49.993 &	 61.832 \\
\hline
\end{tabular}

}
\caption{Mean KGE, NSE and RMSE of each imputation method at varying levels of missingness across the ten stations.}
\label{tab:results}
\end{table}

\begin{figure}
    \centering
    \includegraphics[width = 0.5\textwidth]{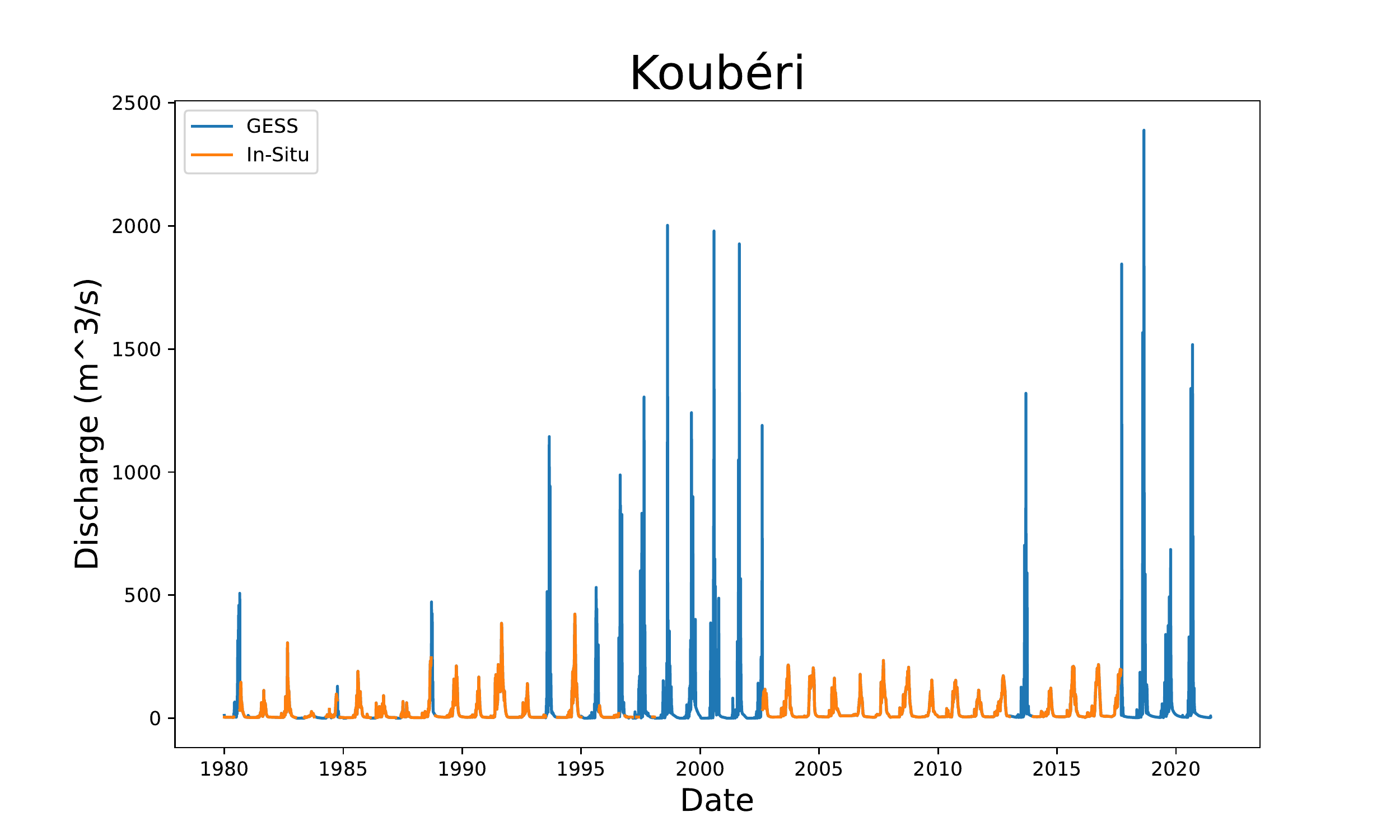}
    \caption{The infilled streamflow time series at the Koubéri station using the GESS lookup Bias Correction.}
    \label{fig:gess_kou}
\end{figure}
\par Table \ref{tab:stream_20} shows the KGE values at each of the stations at a level 20\% missingness. It can be seen that GESS yields the best performance at Yankin and Kompongou stations. Yankin and Kompongou had the highest prevalence of missing data at 47.67\% and 59.48\%, respectively. A possible reason for this result is that, given the elevated levels of missingness at these stations, the complete data could not sufficiently calibrate the bias correction. GP and Elastic Net imputation methods provide the best performance in eight of the ten stations with the highest KGE values obtained at stations with highly complete data in the Athiémè and Bonou stations. QM as bias correction yields KGEs superior to multiple imputations by k-NN and RF, suggesting that bias correction would remain the most efficient approach when constrained by computational resources.

\begin{table*}
    \centering
\scalebox{1}{    
\begin{tabular}{lcccccccccc}
 
\toprule
&\multicolumn{10}{c}{Stations} \\\cline{2-11}\\

Method & Yankin &Lanta &	Kompongou&	Athiémè&	Kaboua &Beterou & Bonou&Porga	& Koubéri&Domè\\
\midrule
KNN	& 0.786 &	 0.728 &	 0.725 &	 0.864 &	 0.772 &	 0.847 &	 0.846& 	 0.851& 	 0.784& 	 0.797\\ 
RF & 0.821 &	 0.784 &	 0.780 &	 0.859& 	 0.841& 	 0.807& 	 0.832 &	 0.839& 	 0.626& 	 0.822 \\

Gaussian Process&	 0.863 &	 0.857& 	 0.829& 	 0.919& 	 0.833 	& 0.831& 	 0.877& 	 \textbf{0.913}& 	\textbf{ 0.914}& 	 0.894\\ 
GESS&	 \textbf{0.896} &	 0.651& 	\textbf{ 0.905}& 	 0.643& 	 0.870& 	 0.865& 	 0.608& 	 0.853& 	 0.443& 	 0.448\\
Elastic Net&	 0.848& 	 \textbf{0.872} &	 0.809& 	 \textbf{0.930} &	\textbf{ 0.881}& 	 \textbf{0.881}& 	 \textbf{0.908} &	 0.894& 	 0.901& 	 0.905\\ 
Quantile Mapping&	 0.831& 	 0.871& 	 0.823& 	 0.936& 	 0.841& 	 0.838& 	 0.906& 	 0.872& 	 0.865 &	 \textbf{0.912}\\ 
\hline
\end{tabular}
}
\caption{KGE of each imputation method at 20\%  missingness at each respective gauging station}
\label{tab:stream_20}
\end{table*}

\begin{figure}
    \centering
    \includegraphics[width = 0.5\textwidth]{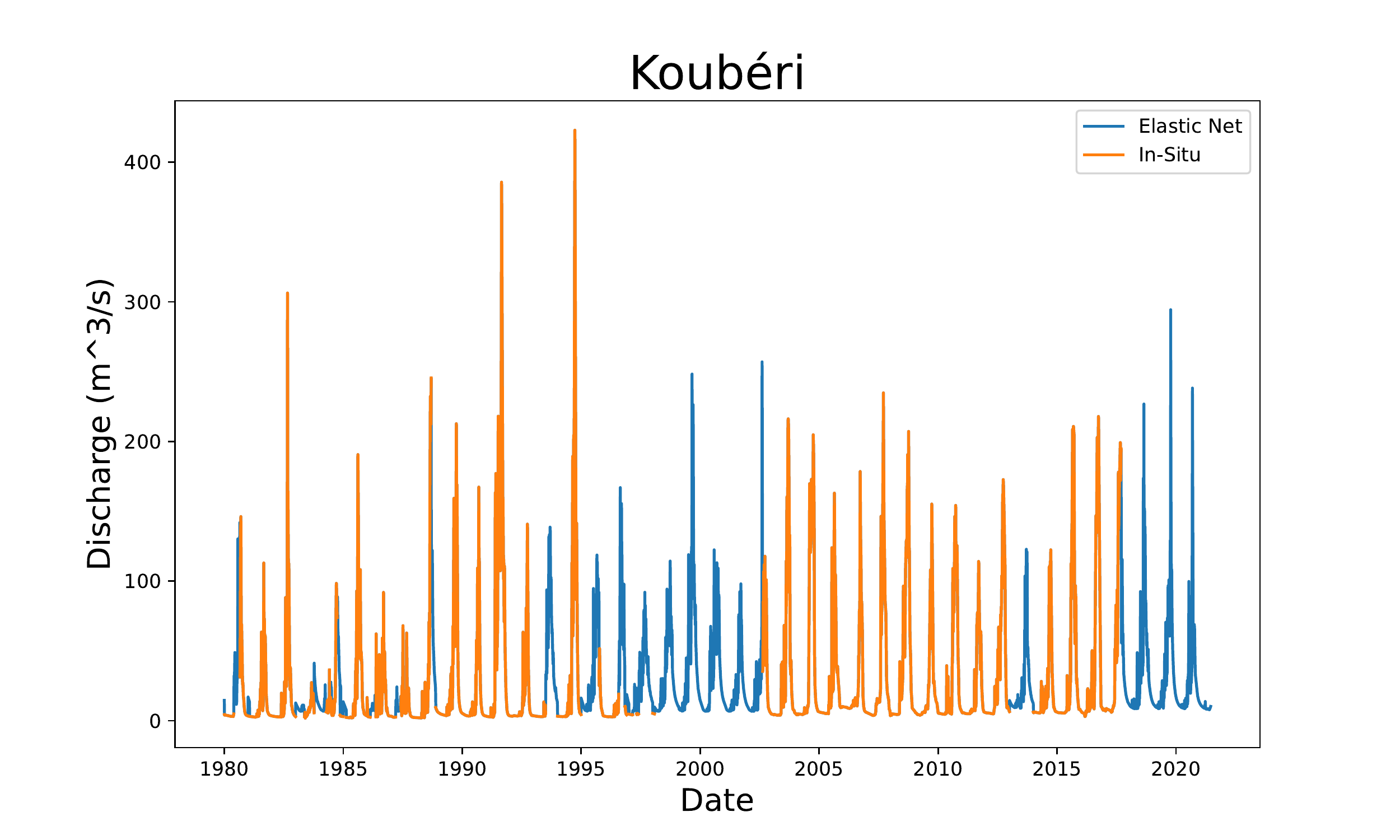}
    \caption{The infilled streamflow time series at the Koubéri station using the Elastic Net Bias Correction. }
    \label{fig:Enet_kou}
\end{figure}

Figure \ref{fig:gess_kou} depicts the embedded bias in the GESS forecasts at the Koubéri station. Figure \ref{fig:gess_kou} shows the resultant time series by looking up the missing data from the GESS. The positive bias in the GESS manifest in that GESS lookups lead to new discharge extremes that are five times the extremes recorded in the in-situ observations. In operational settings, such significant positive biases can lead to false flood alerts and sub-optimal use of already limited resources. The remedy to this extreme bias is seen in Figure \ref{fig:Enet_kou}, where bias correction removes the significant positive bias while retaining the seasonal periodicity in discharge.   

\section{Conclusion}
We evaluated state-of-the-art imputation methods for streamflow prediction, demonstrating the need and importance for bias-correcting GEOGloWS ECMWF streamflow service (GESS) streamflow forecast information. Our findings show that the bias introduced by GESS forecasts can be significant and result in possible false flooding alerts. We minimize such bias using  QM, GP, and Elastic Net regressions trained in periods where {\it in-situ} observations are available. The resultant imputation under simulated missingness shows that GESS bias correction outperforms train k-NN and RF imputation trained on available data alone.

Our findings enhance decision-making based on streamflow model-based forecasts, reducing bias introduced by missing {\it in-situ} data. The reliance on such forecasts is high, particularly in areas where data collection is too costly, under-resourced, or not possible across Sub-Saharan Africa. The proposed bias correction methodology integrates into existing operational GESS forecasting and now enables real-time flooding and drought monitoring with lead times of up to 15 days. Continued work aims to enhance and expand on the operational implementation of our bias-correction-based methodology. Specifically, we aim to interpolate and extrapolate the best-performing bias-correction model to ungauged areas. This is anticipated to vastly increase the utility of GESS to large areas across Sub-Saharan Africa and Central America that are currently ungauged. The robustness of our proposed approach will also be further tested at the seasonal scale to ensure that wet seasons and overall seasonal dynamics are captured accurately. 
\FloatBarrier


\bibliography{aaai22}

\end{document}